\documentclass[letterpaper,10pt,journal,twoside]{IEEEtran}

\usepackage{graphicx}
\usepackage{tikz}
\usetikzlibrary{calc,positioning}

\usepackage{array}
\usepackage{booktabs}
\usepackage{multirow}
\usepackage{tabularx}
\usepackage{diagbox}
\usepackage{adjustbox}
\usepackage{colortbl}

\usepackage{amsmath}
\usepackage{amssymb}
\usepackage{bbm}
\usepackage{dsfont}
\usepackage{siunitx}

\usepackage{subcaption}
\usepackage[font=small]{caption}

\usepackage{algorithm}
\usepackage{algpseudocode}

\usepackage{xcolor}
\usepackage{xspace}
\usepackage{textcomp}
\usepackage{soul}
\usepackage{balance}
\usepackage{romannum}
\usepackage{xparse}
\usepackage{etoolbox}
\usepackage{lipsum}

\usepackage{rpm_acronyms}
\usepackage{rpm_math}
\usepackage{rpm_misc}

\usepackage[hidelinks]{hyperref}
\usepackage{cleveref}

\crefname{figure}{Fig.}{Figs.}
\Crefname{figure}{Fig.}{Figs.}

\makeatletter
\renewcommand{\section}{\@startsection{section}{1}{\z@}%
  {0.8\baselineskip plus 0.2\baselineskip minus 0.1\baselineskip}%
  {0.35\baselineskip plus 0.1\baselineskip minus 0.05\baselineskip}%
  {\centering\normalfont\normalsize\scshape}}

\renewcommand{\subsection}{\@startsection{subsection}{2}{\z@}%
  {0.6\baselineskip plus 0.15\baselineskip minus 0.1\baselineskip}%
  {0.25\baselineskip plus 0.05\baselineskip minus 0.05\baselineskip}%
  {\normalfont\normalsize\itshape}}
\makeatother

\newcommand{\DATASETNAME}{ScaRF}
\newcommand{\METHODNAME}{ScaRF-SLAM}

\definecolor{linkcolor}{HTML}{C46A00}
\definecolor{gray}{RGB}{128,128,128}

\newcommand{\best}[1]{\textbf{#1}}
\newcommand{\gray}[1]{\textcolor{gray}{#1}}

\newcommand{\cmtt}[1]{{\fontfamily{cmtt}\selectfont #1}}

\begin{document}

\title{\METHODNAME:~Scale-Consistent~Reconstruction~with Feed-Forward Models and Classical Visual SLAM}

\author{Yuhao Zhang, Yifu Tao, Frank Dellaert, and Maurice Fallon%
\thanks{Manuscript received: May 28, 2026; Revised: July 28, 2026; Accepted: August 25, 2026.
This paper was recommended for publication by Editor Javier Civera upon evaluation of the Associate Editor and Reviewers’ comments.
This work was partly funded by the National Research Foundation of Korea (NRF) grant funded by the Korea government (MSIT)(No. RS-2024-00461409) and by the UKRI Project Mobile Robotic Inspector (EP/Z531212/1). For the purpose of open access, the authors have applied a Creative Commons Attribution (CC BY) license to any Accepted Manuscript version arising. \textit{(Corresponding author: Yuhao Zhang.)}}%
\thanks{Yuhao Zhang, Yifu Tao, and Maurice Fallon are with Oxford Robotics Institute, University of Oxford, U.K. (e-mail: \cmtt{yuhao@robots.ox.ac.uk}).}%
\thanks{Frank Dellaert is with School of Interactive Computing, Georgia Institute of Technology, USA (e-mail: \cmtt{frank.dellaert@cc.gatech.edu}).}%
}

\markboth{IEEE Robotics and Automation Letters. Preprint Version. Accepted August, 2026}%
{Zhang \MakeLowercase{\textit{et al.}}: ScaRF-SLAM: Scale-Consistent Reconstruction with Feed-Forward Models and Classical Visual SLAM}

\IEEEpubid{~\copyright~2026 IEEE. Digital Object Identifier: 10.1109/LRA.2026.3732836}

\maketitle

\bstctlcite{IEEEexample:BSTcontrol}

\begin{abstract}
Recent works have explored unifying SLAM with geometric foundation models (GFMs). However, directly using GFM predictions for tracking is highly sensitive to model capability and uncertainty, as geometric inaccuracies in the predictions can adversely affect pose estimation. To address this limitation, we propose a decoupled framework that integrates classical feature-based SLAM with GFMs, which achieves higher quality and more consistent dense reconstruction. In brief, we use classical visual SLAM for robust low-latency tracking and use GFMs exclusively for mapping. By anchoring mapping to poses produced by the SLAM module and optimizing across depth scales, the proposed design avoids propagating inaccuracies from GFM predictions into pose estimation while imposing geometric constraints on the reconstruction. The system builds submaps from multiple posed keyframes and enforces scale consistency via lightweight frame and submap scale optimization. It also performs projection-based point cloud fusion within each submap, and updates submaps online to reflect trajectory updates from the feature-based SLAM. To evaluate tracking and reconstruction of our method, we introduce a loop-rich, building-scale indoor dataset with accurate sensor trajectories and LiDAR ground-truth. Experiments show that our approach achieves superior trajectory accuracy while improving reconstruction precision by 10\%--20\% over existing methods, with about 2 cm reconstruction error per 10 m chunk on building-scale dataset. On large-scale outdoor datasets, it attains 10 cm error per 30 m chunk (w.r.t LiDAR ground-truth models).

Code and dataset: \href{https://github.com/ori-drs/ScaRF-SLAM}{\textcolor{linkcolor}{\texttt{github.com/ori-drs/ScaRF-SLAM}}}
\end{abstract}

\begin{IEEEkeywords}
SLAM, Deep Learning for Visual Perception.
\end{IEEEkeywords}

\section{Introduction}
\label{sec:introduction}

\IEEEPARstart{S}{imultaneous} Localization and Mapping (SLAM) has been extensively studied in robotics and has reached a high level of maturity. Classical visual SLAM systems can provide accurate and robust state estimation, while supporting diverse sensing configurations~\cite{handbook}. These systems are highly optimized and reliable in real-world deployments. However, they typically produce sparse or semi-dense maps; high-quality dense reconstruction often requires active depth sensors.


More recently, advances in feed-forward geometric foundation models (GFMs) have enabled the prediction of dense geometry from images. Models such as DepthAnything3~\cite{da3} leverage large-scale training and transformer-based architectures to infer consistent multi-view depth, offering a promising alternative to purely geometric methods.
A key innovation in DA3 and other similar models is the capacity to take camera intrinsics and poses as auxiliary inputs. This enables more explicit geometric conditioning and consistent reconstruction.

\IEEEpubidadjcol
\begin{figure}[t] 
    \centering

    \begin{tikzpicture}
        \node[anchor=south west, inner sep=0] (img) at (0,0)
        {\includegraphics[width=\columnwidth]{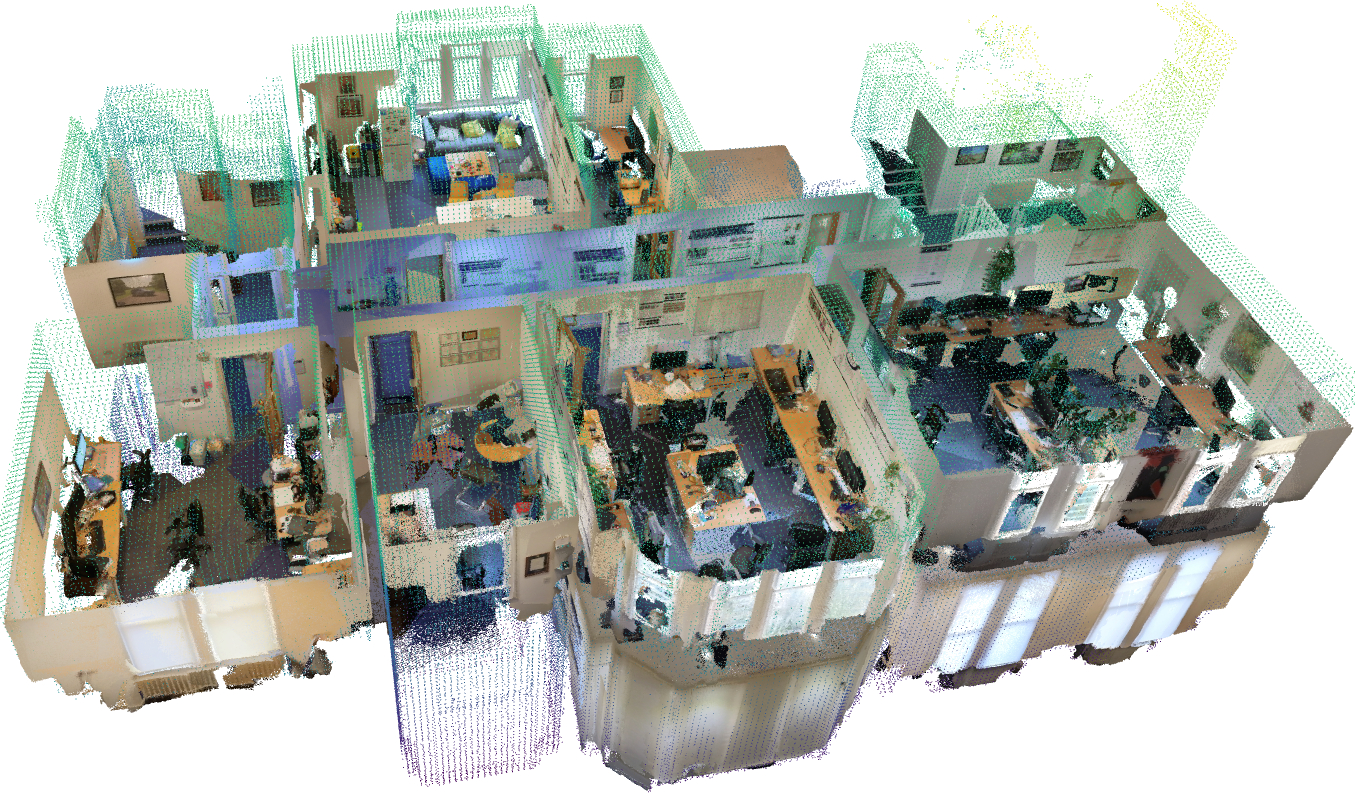}};

        \begin{scope}[x={(img.south east)}, y={(img.north west)}]
            \node[
                anchor=north west,
                text=green!50!black,
            ]
            at (0.00,0.99) {\scriptsize LiDAR Scans →};
        \end{scope}
        \node[
            anchor=south east,
            inner sep=0
        ] (device) at ([xshift=-3pt, yshift=0pt]img.south east)
        {\includegraphics[width=0.1\columnwidth]{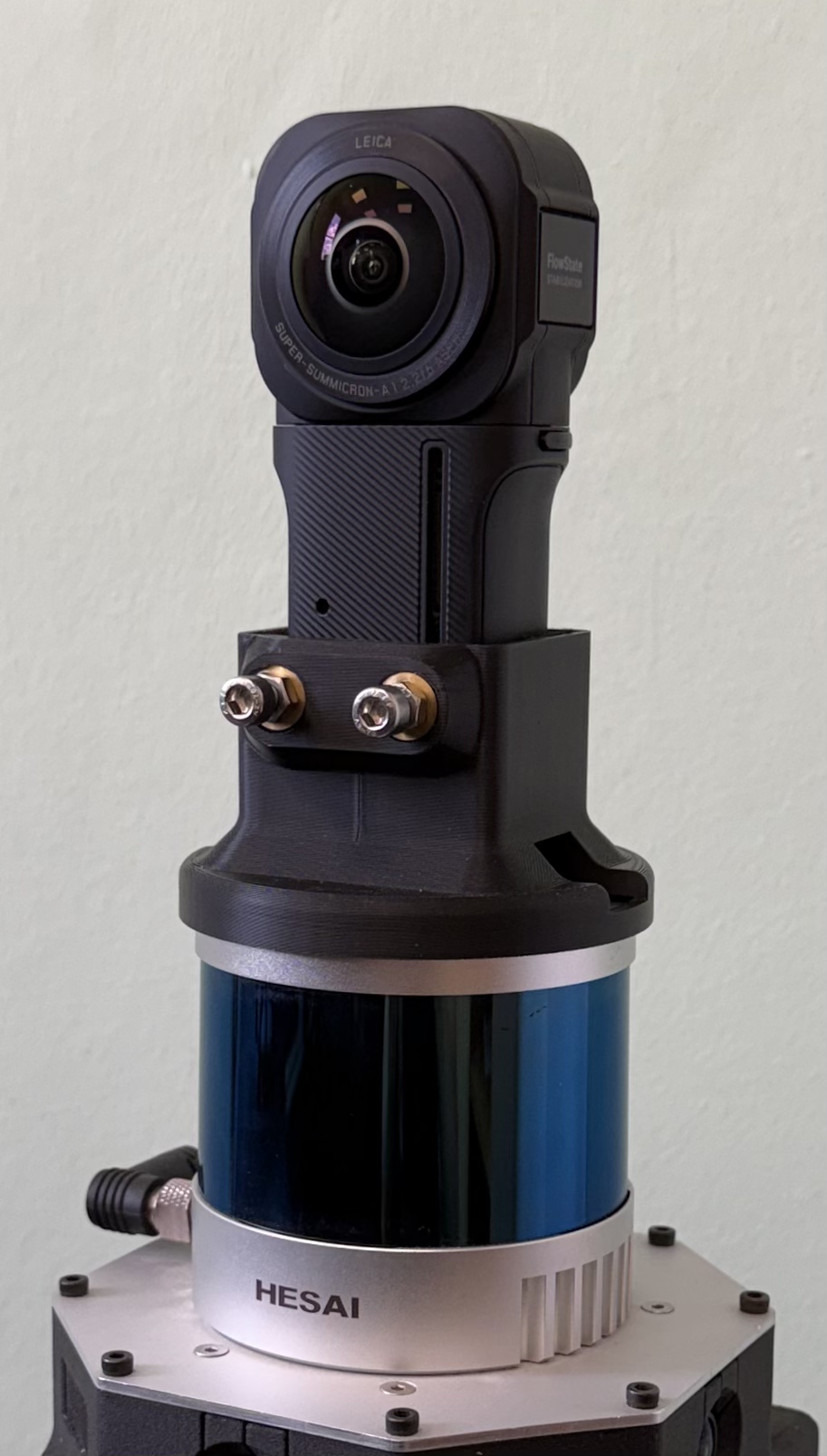}};
        
        \node[
            anchor=south east,
            text=black,
            font=\scriptsize
        ] at ([xshift=-1pt]device.south west)
        {Dataset Device →};
    \end{tikzpicture}

    \vspace{3pt}
    
    \includegraphics[width=\columnwidth]{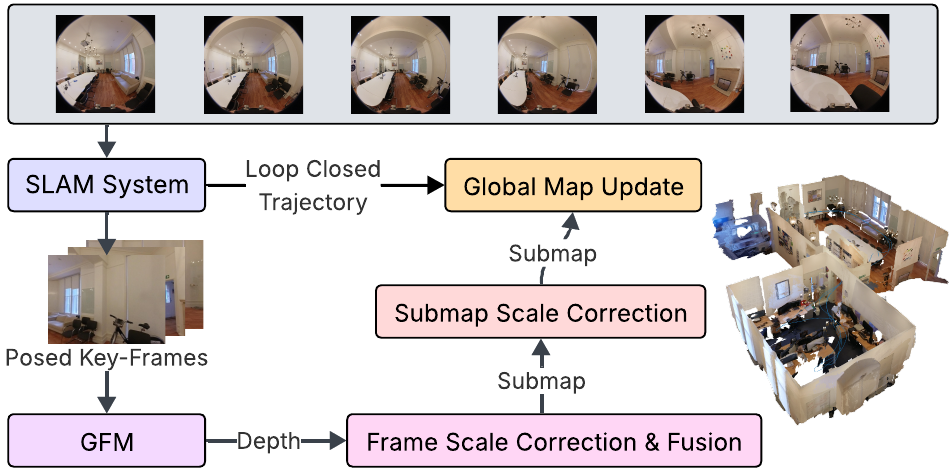}

    \caption{\textit{Top}: A two-floor dense reconstruction produced by our mapping module using ground-truth poses to demonstrate best achievable reconstruction quality. The map is aligned with registered LiDAR scans for visual comparison. \textit{Bottom}: The reconstruction pipeline.} 
    \label{fig:pipeline}
\end{figure}

Despite recent progress, developing performant SLAM systems with GFMs remains challenging. Most existing approaches tightly couple tracking and mapping within a unified GFM-based pipeline~\cite{mast3r-slam, mast3r-fusion, vggt-slam2}. Such designs are inherently sensitive to the quality of GFM predictions, and we observe that prediction quality is influenced by the number of input views and their viewpoint diversity: fewer input views typically degrade geometric accuracy (Tables~\ref{tab:ori_ablation_6} and~\ref{tab:ori_ablation_12}), whereas more views lead to greater GPU memory usage and latency.

Although some fully GFM-based SLAM methods report improved tracking over classical ones (e.g., monocular ORB-SLAM3) on certain benchmarks~\cite{tum-rgbd, eth3d}, we attribute this to dataset bias: these datasets are small-scale and lack proper initialization, which is critical for classical monocular methods. We show that when evaluated on datasets with proper initialization (e.g., EuRoC~\cite{euroc}), classical visual SLAM still achieves higher tracking accuracy (Section~\ref{sec:mono_tracking}). Furthermore, incorporating additional sensing modality~\cite{mast3r-fusion} or more advanced camera configurations (e.g., fisheye, multi-camera) into these tightly coupled GFM-based methods requires development of alternative architectures. This makes it difficult to leverage the maturity and modularity of existing SLAM frameworks.

In this work, we take a different approach and introduce \METHODNAME\ (\underline{Sca}le-Consistent \underline{R}econstruction with \underline{F}eed-Forward Models), a decoupled framework that combines classical feature-based SLAM and GFMs in a complementary manner. Instead of relying on GFMs for state estimation, we leverage the maturity of classical SLAM that supports diverse modalities (e.g., visual-inertial), camera rigs (e.g., multi-camera), and camera models (e.g., fisheye) for tracking, while using GFMs for dense mapping only. The key idea is to anchor the mapping process using the SLAM poses and optimize the reconstruction on top of them. This design avoids inaccurate GFM predictions propagating into tracking errors, while still benefiting from the dense GFM reconstructions.

Our system operates by grouping posed keyframes into small batches and feeding them into a GFM for depth prediction to construct submaps. We then enforce scale consistency using a lightweight multi-view optimization at both inter-frame and inter-submap level, and carry out a local point cloud fusion step before updating the global map. Notably, this framework is robust to GFM performance degradation under small input image batch sizes, making it well suited for online and resource constrained scenarios (Table~\ref{tab:ori_ablation_6}).

The main contributions are summarized as follows:
\begin{itemize}
    \item A practical decoupled GFM-based mapping framework that integrates naturally with existing feature-based visual SLAM systems. It preserves the accurate, low-latency pose estimation and sensor reconfigurability of classical SLAM systems, while fully leveraging the intrinsic and pose input modes of recent GFMs to achieve high-quality dense reconstruction from passive vision.
    \item An efficient scale optimization and point cloud fusion mechanism that achieves globally consistent reconstruction for sequences of any length and is robust to GFM performance degradation under small input batch sizes.
\end{itemize}

We extensively evaluate the system on public datasets (outdoor), high-quality self-collected dataset with accurate ground-truth trajectories and point clouds (indoor, building-scale), and multi-session data collected by a robotic platform. Our code and dataset are released for future studies to evaluate against.
\section{Related Work}
\label{sec:related_work}

\subsection{Feature-based SLAM}
Classical feature-based SLAM systems estimate camera poses by tracking sparse keypoints and enforcing geometric consistency over time~\cite{handbook}. Odometry systems such as OpenVINS~\cite{openvins} fuse visual features with inertial measurements for efficient state estimation. ORB-SLAM3~\cite{orb-slam3} integrates tracking, local mapping, and loop closure within a keyframe-based framework and supports diverse sensor configurations. Maplab~\cite{maplab} is a framework for large-scale, multi-session, and multi-robot mapping with collaborative workflows. These systems demonstrate the maturity and flexibility of feature-based SLAM, but they do not create dense 3D reconstructions.

\subsection{Dense SLAM and Mapping}
Dense SLAM methods aim to reconstruct full scene geometry. Many systems rely on active depth measurements from range sensors. Early systems such as KinectFusion~\cite{kinectfusion} and ElasticFusion~\cite{elasticfusion} demonstrate real-time dense reconstruction via TSDF or surfel fusion. To relax this dependency, subsequent works pursue monocular dense reconstruction using learned depth. CodeSLAM~\cite{codeslam} introduces a compact learned depth representation jointly optimized with camera poses. While effective, it tightly couples geometry with pose optimization in a latent space, limiting flexibility and making it difficult to adapt to large-scale settings.

More recent works instead leverage SLAM-derived geometric priors to guide depth prediction. SimpleMapping~\cite{simplemapping} uses ORB-SLAM3 poses and sparse landmarks to guide multi-view stereo networks, while OKVIS2-X~\cite{okvis-d, okvis2-x} exploits visual-inertial pose estimates for uncertainty-aware depth fusion and tightly coupled with a factor-graph SLAM backend. These approaches reflect a growing trend of integrating classical SLAM with learned depth before the emergence of GFMs.

While they typically integrate the results into volumetric representations with probabilistic sensor fusion to suppress noise in the estimated depths, our method operates directly on point clouds. It performs lightweight scale optimization constrained by fixed SLAM poses to enforce geometric consistency. This design enables a more memory-efficient and scalable mapping framework while preserving geometric accuracy.

\subsection{GFM-enabled SLAM}
Recently developed GFMs~\cite{mast3r, vggt, mapanything, da3} leverage strong learned priors to estimate dense geometry from multi-view monocular inputs and exhibit strong cross-scene generalization. However, their high GPU memory demands limit the input batch sizes on resource-constrained devices, motivating temporal chunking with $\mathrm{Sim}(3)$ alignment for long sequences~\cite{vggt-long,laser}. Recent work has also explored stateful streaming~\cite{cut3r,long3r,ttt3r,lingbotmap} and scalable offline reconstruction~\cite{scal3r,vggt3}. While these methods can estimate poses and dense geometry over long sequences, they fall short of providing a complete solution to the SLAM problem.

Complementary to these methods, several works incorporate GFMs into full SLAM pipeline. MASt3R-SLAM~\cite{mast3r-slam} builds a real-time dense SLAM system upon the two-view 3D reconstruction priors, using dense pointmap predictions and learned correspondences for {camera tracking}, {relocalization}, {loop closure}, and {$\mathrm{Sim}(3)$ optimization}. Similarly, ViSTA-SLAM~\cite{vista-slam} uses a symmetric two-view model within a pose graph framework. AIM-SLAM~\cite{aim-slam} moves beyond fixed-length inputs by adaptively selecting a variable-size set of keyframes for prediction. Submap-based approaches such as VGGT-SLAM 2.0~\cite{vggt-slam2} construct local multi-view submaps from GFM predictions and align them via intra- and inter-submap optimization. MASt3R-Fusion~\cite{mast3r-fusion} further extends this line of work by incorporating IMU and GNSS.

Despite promising performance, integrating GFMs into SLAM pipelines typically entails substantial engineering effort, particularly when extending to auxiliary sensing modalities~\cite{mast3r-fusion}, and remains challenging in more advanced settings such as multi-session operation, multi-camera rigs, and fisheye cameras. Moreover, these methods rely on aligning dense predictions that are not strictly geometrically accurate and consistent, requiring high-dimensional optimization. In contrast, our method adopts a decoupled design that leverages GFMs solely for mapping, simplifying map optimization to scale correction and enforcing geometric constraints derived from mature feature-based SLAM upon the learned geometry.

\section{Methodology}
\label{sec:methodology}

An overview of the proposed framework is shown in Fig.~\ref{fig:pipeline}. Input images are processed by a classical SLAM system, which provides instantaneous pose estimates. We group recent posed keyframes into a batch, rectify them to pinhole images, and feed to a GFM for depth prediction. Each batch defines a submap. We enforce scale consistency via frame-level optimization within each submap and submap-level optimization between submaps, both anchored to the SLAM poses. The map is also updated online to reflect trajectory changes in the SLAM system caused by loop closures.

Our decoupled design offers three key advantages.
First, anchoring dense mapping to accurate SLAM poses enables the framework to achieve globally consistent reconstruction even when GFM performance degrades at small input batch sizes (Section~\ref{sec:recon_eval}). Unlike methods that optimize $\mathrm{Sim}(3)$ alignment on GFM predictions, our method is fully anchored to the SLAM poses and only optimizes the scale factors. This is motivated by the observation that GFM predictions are not fully geometrically accurate, even up to scale~\cite{laser}; thus, poses obtained by $\mathrm{Sim}(3)$ optimization over these predictions are highly sensitive to the performance of GFM.

Second, it avoids the reconstruction quality and tracking latency trade-off inherent to fully GFM-based SLAM: larger input batches improve prediction quality through greater viewpoint diversity but increase latency, making integration with robotic systems less practical. In contrast, our framework preserves high-frequency, low-latency tracking with feature-based SLAM, while only dense mapping runs asynchronously at slightly higher latency. Moreover, the SLAM poses also provide strong priors that can improve GFM predictions when exploiting the pose input option of recent models~\cite{da3}.

Third, it naturally leverages the maturity of classical SLAM systems, which achieve accurate pose estimation (Section~\ref{sec:mono_tracking}) and support diverse sensing modalities (e.g., visual-inertial), camera rigs (e.g., monocular, stereo, multi-camera), and camera models (e.g., fisheye). In particular, classical SLAM systems can use wide-FoV fisheye cameras to achieve robust tracking in confined spaces, yet current GFMs are primarily trained on rectified pinhole images with limited FoV.

\subsection{Notation and Preliminary}
\label{sec:methodology:notation}
In this paper, matrices and vectors are denoted by bold symbols (e.g., $\mathbf{I}$), while sets are represented in calligraphic font (e.g., $\mathcal{I}$).
A submap consists of a set of frames $\mathcal{I}$, with associated image data $\mathbf{I}$, intrinsics $\mathbf{K}$, poses $\mathbf{T}$, depth maps $\mathbf{D}$, and confidence maps $\mathbf{C}$. We use subscripts to denote frame indices and superscripts to denote submap indices; i.e., $\mathcal{I}_i^j$ denotes the $i$-th frame of the $j$-th submap $\mathcal{S}^j$.

We adopt DepthAnything3 (DA3)~\cite{da3} as our geometric foundation model.
Given a batch of images $\mathbf{I}$ from frames $\mathcal{I}$, along with optional camera intrinsics $\mathbf{K}$ and poses $\mathbf{T}$, DA3 predicts per-frame dense depth maps $\mathbf{D}$ and confidence maps $\mathbf{C}$. When camera poses are provided, DA3 aligns its internally predicted poses to the input $\mathbf{T}$ via a $\mathrm{Sim}(3)$ transformation and scales $\mathbf{D}$ accordingly, then explicitly overwrites the predicted intrinsics and poses with the input $\mathbf{K}$ and $\mathbf{T}$.

\subsection{Frame-Level Scale Optimization}
\label{sec:methodology:frame_scale}
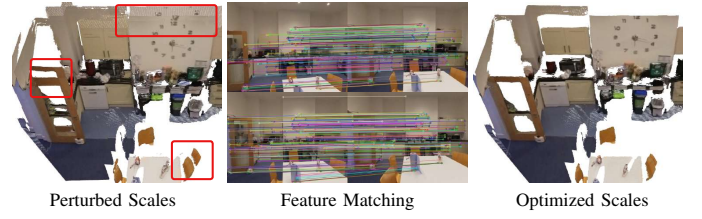
\begin{figure}[t!]
\centering
\begin{tikzpicture}
    \node[anchor=south west, inner sep=0] (img) at (0,0)
        {\includegraphics[width=\columnwidth]{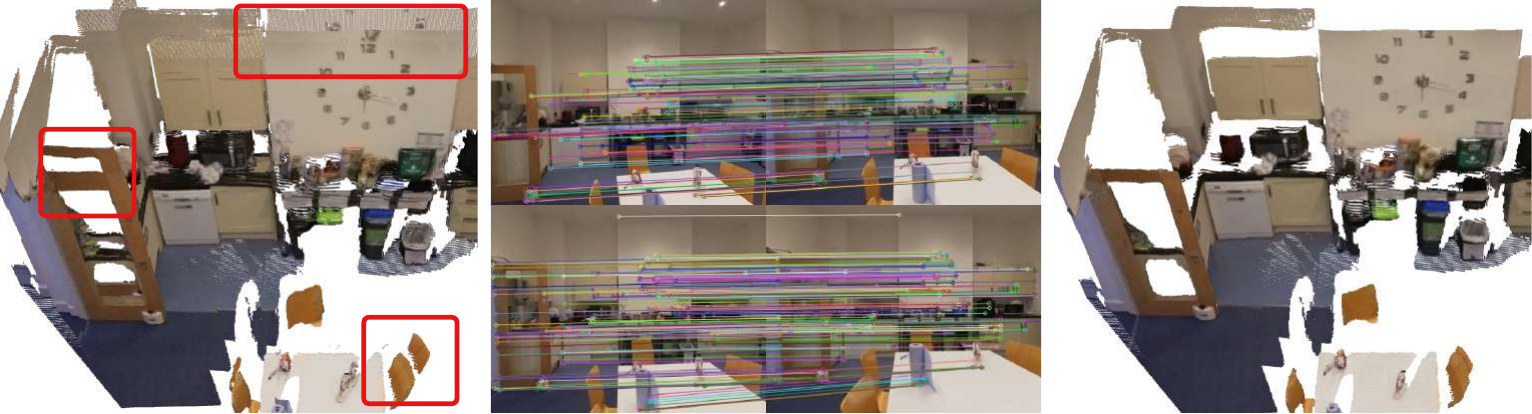}};

    \begin{scope}[x={(img.south east)}, y={(img.north west)}]
        \node[
            anchor=north,
            font=\scriptsize,
            inner sep=2pt
        ] at (0.15,-0.02) {Perturbed Scales};
        \node[
            anchor=north,
            font=\scriptsize,
            inner sep=2pt
        ] at (0.50,-0.02) {Feature Matching};
        \node[
            anchor=north,
            font=\scriptsize,
            inner sep=2pt
        ] at (0.85,-0.02) {Optimized Scales};
    \end{scope}
\end{tikzpicture}
\caption{Frame-scale optimization. For illustration purposes, depth scales are randomly perturbed before optimization. After feature matching, the optimization yields consistent scales across frames.}
\label{fig:frm_opt}
\end{figure}

Once DA3 predicts depth maps $\mathbf{D}$ and confidence maps $\mathbf{C}$ for frames $\mathcal{I}$, we first discard depth values with confidence below a threshold and back-project the remaining pixels to obtain a point cloud $\mathbf{P}$. Although $\mathbf{P}$ may visually appear to be geometrically consistent due to the capability of DA3, it is more properly described as a prior than a measurement.

To achieve better scale consistency within each submap, we introduce a per-frame scale variable $s_i \in \mathbb{R}^{+}$ for each frame $\mathcal{I}_i$, and optimize these variables using sparse feature correspondences. For each image pair $(\mathcal{I}_i, \mathcal{I}_j)$, we obtain sparse matches $\mathcal{M}_{ij}$ using LightGlue~\cite{lightglue} with geometric verification, then we back-project matched pixels into 3D points via $\pi^{-1}(\cdot; \mathbf{K}_i)$. The scale variables are then estimated by minimizing the L2 distance between matched 3D points across all frames of the submap using GTSAM~\cite{gtsam}:
\begin{align*}
\min_{\{s_i\}} \sum_{(i,j)} \sum_{(\mathbf{u}_i, \mathbf{u}_j) \in \mathcal{M}_{ij}} 
\Big\| \;
& \mathbf{T}_i \big( s_i \, \pi^{-1}(\mathbf{u}_i, \mathbf{D}_i(\mathbf{u}_i); \mathbf{K}_i) \big) \nonumber \\
- \; & \mathbf{T}_j \big( s_j \, \pi^{-1}(\mathbf{u}_j, \mathbf{D}_j(\mathbf{u}_j); \mathbf{K}_j) \big)
\;\Big\|_2^2,
\end{align*}
where $\mathbf{T}_i$ and $\mathbf{T}_j$ denote the poses of frames $\mathcal{I}_i$ and $\mathcal{I}_j$, respectively, as given by the underlying SLAM system. A visualization is shown in Fig.~\ref{fig:frm_opt}.

\subsection{Point Cloud Fusion}
\label{sec:methodology:fusion}
\newlength{\subimagewidth}
\setlength{\subimagewidth}{0.36\columnwidth} 

\newlength{\arrowshift}
\setlength{\arrowshift}{-5pt} 

\begin{figure}[t]
    \centering

    \makebox[\columnwidth][c]{%
        \begin{minipage}[c]{\subimagewidth}
            \centering
            \includegraphics[width=\linewidth]
                {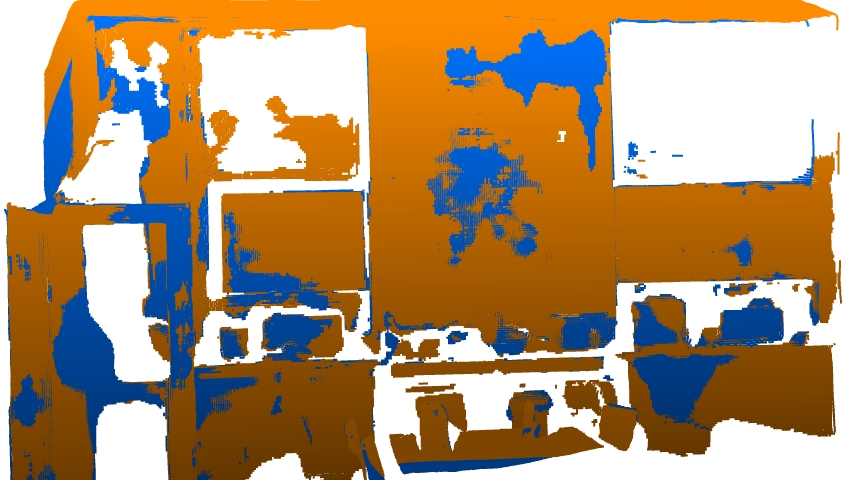}
        \end{minipage}%
        \hfill
        \makebox[0pt][c]{%
            \raisebox{\arrowshift}{%
                $\xrightarrow{\substack{
                    \text{\#points reduced}\\[-1pt]
                    \text{by 41\%}
                }}$
            }%
        }%
        \hfill
        \begin{minipage}[c]{\subimagewidth}
            \centering
            \includegraphics[width=\linewidth]
                {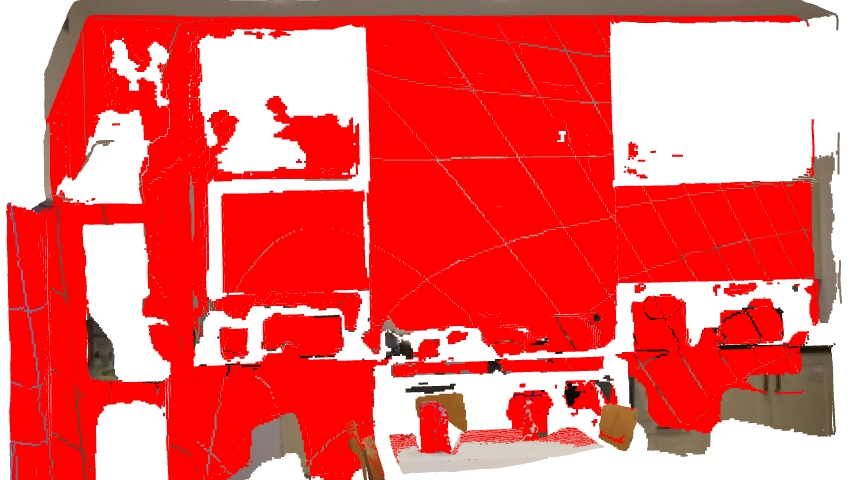}
        \end{minipage}%
    }

    \caption{Point cloud fusion. \textit{Left}: predictions from two
    frames, shown in different colors. \textit{Right}: red points
    indicate fused points.}
    \label{fig:pts_fus}
\end{figure}

To maintain a lightweight map and reduce noise, we perform a point cloud fusion step within the submap after frame-level scale optimization. Unlike MASt3R-SLAM, which performs pointmap fusion based on the per-pixel matching feature produced by MASt3R, DA3 predicts geometry directly without providing matching features. We therefore assume that dense correspondences are implicitly encoded in the predicted geometry, and we can derive matches directly from the geometry that are post-refined by the frame-level scale optimization.

Specifically, for each frame $\mathcal{I}_i$ with point cloud $\mathbf{P}_i$, we project its points into a reference frame $\mathcal{I}_j$ using the camera intrinsics $\mathbf{K}_j$ and poses $\mathbf{T}_i, \mathbf{T}_j$. A correspondence is established if the projected depth is within a threshold of the observed depth. Matched points are then fused via confidence-weighted averaging, with weights derived from $\mathbf{C}_i$ and $\mathbf{C}_j$ (Fig.~\ref{fig:pts_fus}).

\subsection{Submap-Level Scale Optimization}
After frame-level scale optimization and point cloud fusion, each submap becomes locally consistent. We anchor the point cloud $\mathbf{P}$ of each submap to the pose of its central frame $\mathbf{T}_i$. Similar to~\cite{vggt-slam, vggt-slam2}, consecutive submaps are constructed with one overlapping frame. For two consecutive submaps $\mathcal{S}^i$ and $\mathcal{S}^{i+1}$, the last frame image of $\mathcal{S}^i$ coincides with the first frame image of $\mathcal{S}^{i+1}$, i.e., $\mathbf{I}_{-1}^i = \mathbf{I}_{0}^{i+1}$. This overlap naturally provides matched points between submaps.

To enforce global scale consistency across submaps, we introduce a per-submap scale variable $s^i \in \mathbb{R}^{+}$ for each submap $\mathcal{S}^i$. Given a set of matched points pairs $\mathcal{M}^{i,i+1}$, we optimize the scale factors by minimizing the L2 distance between corresponding 3D points:
\begin{align*}
\min_{\{s^j\}} \sum_{(j,j+1)} \sum_{(\mathbf{p}_i, \mathbf{p}_{i'}) \in \mathcal{M}^{j,j+1}}
\Big\| \mathbf{T}^j \big( s^j \mathbf{p}_i \big) \nonumber
- \mathbf{T}^{j+1} \big( s^{j+1} \mathbf{p}_{i'} \big)
\;\Big\|_2^2
\end{align*}
where $s^j$ denotes the scale of submap $\mathcal{S}^j$, and $\mathbf{T}^j$ denotes the pose of the submap. Compared to frame-level optimization, larger inter-submap baselines improve scale observability and provide stronger geometric constraints.

For efficiency, we use a sliding-window strategy in which only the latest $N$ submaps are optimized when a new submap is created. Global submap optimization is performed once the SLAM system updates the trajectory (i.e., after loop closure).

\subsection{Map Update}
The mapping module is developed to be a component of an incremental SLAM system, rather than being a post-processing method. Thus, we need to update the live map according to the trajectory maintained by the SLAM system. When a loop closure is detected and the trajectory is updated, we use LightGlue matches to build edges between submaps containing the frames involved in the loop closure, transform each submap using the updated pose to which it is anchored, and then invoke submap-level scale optimization with the new poses.

\subsection{Mapping Keyframe Selection}
\label{sec:methodology:keyframe}

\begin{figure}[t] 
    \centering    

    \begin{tikzpicture}
        \node[inner sep=0] (img)
        {\includegraphics[width=0.8\columnwidth]{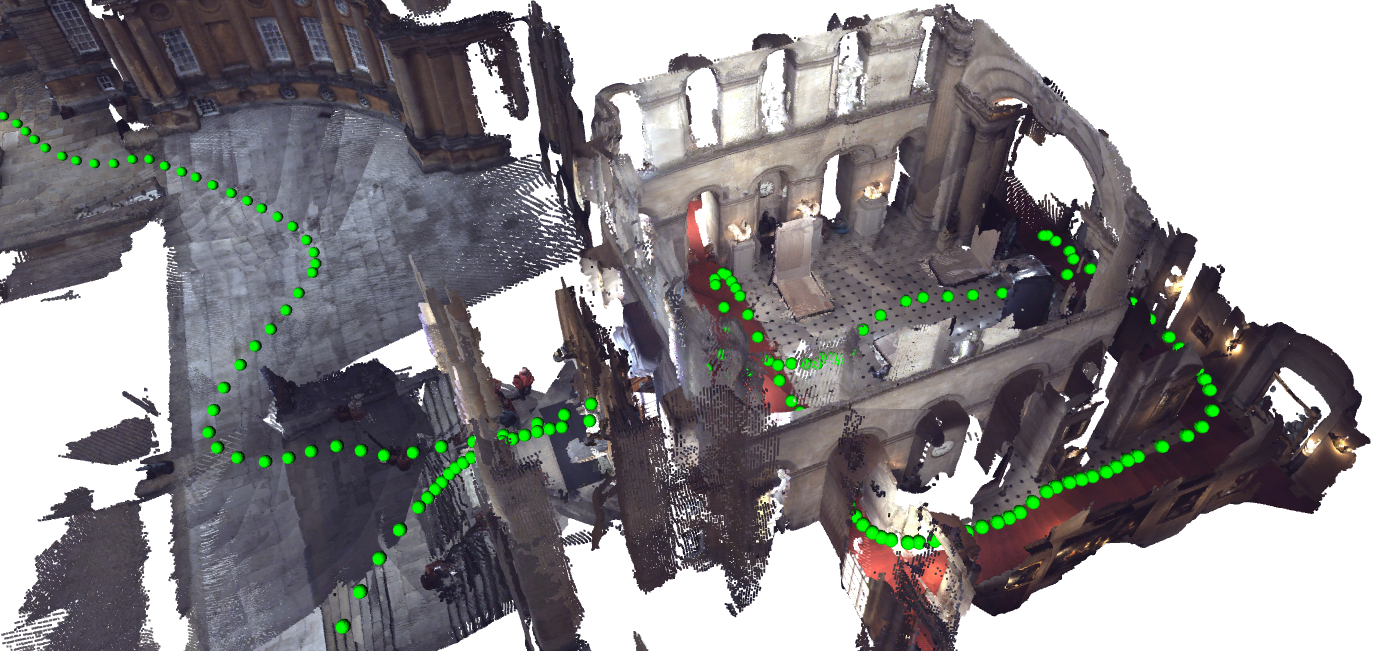}};

        \node[
            overlay,
            text=blue,
            font=\footnotesize\bfseries,
            anchor=north east
        ] at ($(img.north east)+(-0.3,-2.8)$)
        {corridor};

        \node[
            overlay,
            text=lime,
            font=\footnotesize\bfseries,
            anchor=north east
        ] at ($(img.north east)+(-2.5,-1.2)$)
        {hall};

        \node[
            overlay,
            text=lime,
            font=\footnotesize\bfseries,
            anchor=north east
        ] at ($(img.north east)+(-4.8,-0.5)$)
        {outdoor};

    \end{tikzpicture}

    \caption{Visualization of keyframe selection, where the baseline between keyframes is larger in open spaces (outdoor and hall) and smaller in more confined environments (corridor).} 
    \label{fig:kf_selection}
\end{figure}
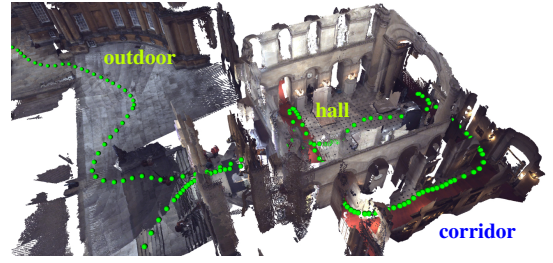

Selecting images from diverse viewpoints is crucial for high-quality GFM prediction with limited image batch sizes. In SLAM systems, such constraints are common due to limited GPU memory and the need for incremental reconstruction. A fixed selection strategy is usually suboptimal: small baselines between frames in open spaces provide insufficient geometric cues for far-range geometry, whereas using a large baseline in near-field scenes reduces visual overlap, resulting in overly sparse keyframes and incomplete reconstruction. Therefore, an adaptive strategy that adjusts baseline according to scene geometry is required.

Given a high-frequency image stream, we first temporally downsample it to construct a candidate set $\mathcal{I}$ whose frequency is aligned with the processing rate of the mapping module (see Section~\ref{sec:evaluation:runtime}). Keyframes are then selected based on relative motion derived from SLAM poses. For a candidate frame $\mathcal{I}_k$ and the last selected keyframe $\mathcal{I}_{k'}$, we compute the positional and angular distance ($\Delta t_{k,k'}$ and $\Delta R_{k,k'}$). A frame is selected if
$
\Delta t_{k,k'} > \tau_t \; \text{or} \; \Delta R_{k,k'} > \tau_R,
$
with fixed angular threshold $\tau_R$ and adaptive translation threshold $\tau_t$.

To adapt to scene scale, we compute the median depth of the latest frame in the current batch,
$
d_{\mathrm{med}} = \mathrm{median}(\mathbf{D}_{-1}),
$
and adjust $\tau_t$ for the next batch accordingly. Specifically, $\tau_t$ is increased when $d_{\mathrm{med}} > \tau_d$, indicating that next batch is likely to observe a large-scale open scene; otherwise, it is kept small to favor near-field reconstruction.

Additionally, we apply range filtering within each batch by discarding depths exceeding
$
d_{\max} = \alpha \tau_t,
$
with $\alpha=20$, to suppress unreliable far-range estimates. This adaptive strategy balances viewpoint diversity and geometric reliability across varying environments (Fig.~\ref{fig:kf_selection}).

\section{Dataset}
\label{sec:dataset}

Existing indoor SLAM datasets are often limited in scale or lack high-quality ground-truth reconstruction~\cite{tum-rgbd, eth3d, euroc}. Therefore, we introduce the \DATASETNAME\ dataset, featuring a fisheye camera traversing a multi-floor office environment with multiple loops, confined staircases, and low-texture corridors that present challenging conditions for dense visual SLAM.

The dataset is captured using an Insta360 ONE RS 1-Inch camera equipped with dual fisheye cameras (FoV $>180^\circ$) and an IMU. We use only the monocular images from the front-facing camera, as the rear camera observes the operator. To obtain ground-truth geometry, the camera is rigidly mounted to a LiDAR-inertial mapping device (Fig.~\ref{fig:pipeline}). Two devices are temporally aligned via their IMU signals. Ground-truth camera poses are obtained by registering each LiDAR scan to a high-precision terrestrial laser scanner (TLS) map and applying calibrated LiDAR-to-camera extrinsics.
For reconstruction evaluation, we use the registered LiDAR scans as the ground-truth 3D model, since the TLS scan was captured months earlier and does not fully reflect the current environment. We additionally provide a filtered version with points outside the camera FoV removed for recall (completeness) evaluation.

\section{Evaluation}
\label{sec:evaluation}

\begin{table}[t!]
\centering
\scriptsize
\caption{Evaluation of ATE (m) for different methods on the EuRoC dataset. The initialization phase is excluded from evaluation.}

\begin{tabular*}{\columnwidth}{@{\extracolsep{\fill}}lccccc}
\toprule
 & MH01 & MH02 & MH03 & MH04 & MH05 \\
\midrule
DA3-Long
& 0.3481
& 0.5270
& 0.6023
& 0.3333
& 0.5918 \\
MASt3R-SLAM 
& \underline{0.0274} 
& \underline{0.0291} 
& \underline{0.0580} 
& \underline{0.1180} 
& \underline{0.0674} \\
VGGT-SLAM2
& 0.0614
& 0.0800
& 0.1284
& 0.6447
& 0.3560 \\
VGGT(D)-SLAM2
& 0.0962
& 0.0718
& 0.0620
& 0.3456
& 0.2246 \\
ORB-SLAM3$^{*}$
& \best{0.0206}
& \best{0.0193}
& \best{0.0298}
& \best{0.0996}
& \best{0.0459} \\

\bottomrule
\end{tabular*}

\vspace{2pt}
\raggedright
\footnotesize
$^{*}$ Monocular;\,
All methods use calibrated intrinsics and rectified images.
\label{tab:euroc_traj_cmp}
\end{table}

\begin{table}[t!]
\centering
\scriptsize
\caption{Evaluation of ATE (m) on the \DATASETNAME\ dataset.}

\begin{tabular*}{\columnwidth}{@{\extracolsep{\fill}}lccccc}
\toprule
 & R01 & R02 & R03 & R04 & R05 \\
\midrule
DA3-Long$^{*}$ 
& 0.6864
& 0.2435
& 0.2556
& 0.5735
& 0.1065 \\
MASt3R-SLAM$^{*}$ 
& \gray{Failed}
& 0.0954
& 0.0855
& 0.3995
& 0.0909 \\
MASt3R-Fusion$^{\dagger}$ 
& 0.1278
& 0.2496
& 0.1183
& 0.1561
& 0.1403 \\
VGGT-SLAM2$^{*}$
& 1.0573
& 0.9097
& 0.8694
& 0.7667
& 0.4423 \\
VGGT(D)-SLAM2$^{*}$
& 0.5362
& 0.1916
& 0.1088
& 0.2572
& 0.1256 \\
\arrayrulecolor{gray!60}
\midrule
\arrayrulecolor{black}
ORB-SLAM3$^{*}$
& 0.1114
& 0.0930
& 0.0773
& 0.0915
& \underline{0.0496} \\
ORB-SLAM3$^{\dagger}$
& \best{0.0840}
& \underline{0.0786}
& \underline{0.0765}
& \best{0.0587}
& 0.1104 \\
OV-SLAM$^{\dagger}$
& \underline{0.1075}
& \best{0.0444}
& \best{0.0316}
& \underline{0.0644}
& \best{0.0309} \\
\bottomrule
\end{tabular*}

\vspace{3pt}
\raggedright
\footnotesize
$^{*}$ Monocular;\,
$^{\dagger}$ Monocular-Inertial;\,
All methods use calibrated intrinsics, if supported.\,
Only first $100\,\mathrm{s}$ used due to OOM issue with MASt3R-SLAM.

\label{tab:ori_traj_cmp}
\end{table}

In this section, we evaluate both tracking and reconstruction performance across multiple datasets. We first compare the camera tracking performance of existing GFM-enabled methods and classical approaches (Section~\ref{sec:mono_tracking}), showing that properly initialized classical methods still provide more accurate tracking. We then evaluate reconstruction quality on our \DATASETNAME\ dataset. To further assess performance in outdoor environments, we include representative KITTI sequences 00 and 05~\cite{kitti}, which contain frequent loop closures, as well as Blenheim-Palace-01 and Christ-Church-02 from Oxford Spires dataset~\cite{oxford_spires}, which feature indoor-outdoor transitions. Ground-truth point clouds are obtained by transforming undistorted LiDAR scans into world frame using ground-truth poses.

For trajectory evaluation, we adopt \textit{Absolute Trajectory Error (ATE)}. For reconstruction evaluation, we use \textit{precision}, \textit{recall}, and \textit{reconstruction error}. Precision measures the proportion of reconstructed points that lie within a threshold of ground-truth points, while recall measures the proportion of ground-truth points covered by the reconstruction. Reconstruction error is defined as the average distance from reconstructed points to their nearest ground-truth neighbors.
We do not report coverage error (the average distance from ground-truth to reconstructed points), as regions filtered out due to low prediction confidence, a common practice in GFM-based reconstruction, can incur artificially large distances and bias this metric. Instead, recall serves as a proportion-based proxy for completeness, while reconstruction error captures geometric fidelity in metric units.

All evaluated GFM-based methods use rectified pinhole images with calibrated intrinsics. Our system uses the raw fisheye images for pose estimation and the pinhole images for GFM prediction. Experiments are conducted on a laptop with an Intel Core Ultra 7 265HX CPU and an NVIDIA RTX PRO 3000 Blackwell GPU with 12GB of memory.

\subsection{Monocular Tracking}
\label{sec:mono_tracking}

In this section, we show that, with proper initialization, classical methods continue to outperform fully GFM-enabled approaches in both single-modal and multi-modal settings, which motivates our design choice to decouple state estimation from dense reconstruction when leveraging GFMs.

\subsubsection{Monocular SLAM}
Table~\ref{tab:euroc_traj_cmp} compares the state-of-the-art GFM-enabled SLAM systems, MASt3R-SLAM~\cite{mast3r-slam} and VGGT-SLAM2~\cite{vggt-slam2}, as well as DA3-Long (script provided by DA3 for long-sequence processing adapted from VGGT-Long~\cite{vggt-long}), with the classical feature-based method ORB-SLAM3~\cite{orb-slam3}. We also create and evaluate a version of VGGT-SLAM2 using the latest geometric foundation model DA3, denoted as VGGT(D)-SLAM2. 
All methods use rectified pinhole images, including ORB-SLAM3. Trajectories are aligned and rescaled to the ground-truth before evaluation.

The results show that, when properly initialized, ORB-SLAM3 consistently outperforms fully GFM-enabled methods in tracking. We attribute the performance gap to geometric inaccuracies in GFM predictions, which in turn adversely affect pose estimation. Replacing VGGT with DA3 leads to a clear improvement in VGGT(D)-SLAM2, suggesting that fully GFM-enabled SLAM remains a promising direction and that future models may further narrow or close the performance gap with respect to classical methods.

\subsubsection{Mono-Inertial SLAM}
We further evaluate existing methods on \DATASETNAME\ dataset with inertial sensing and fisheye images. We include MASt3R-Fusion and OpenVINS augmented with pose graph optimization (denoted as OV-SLAM). MASt3R-SLAM suffers from CUDA out-of-memory issues, so we limit evaluation to the first $100\,\mathrm{s}$. Aside from the OOM issue, it achieves competitive performance.
Table~\ref{tab:ori_traj_cmp} again shows that classical methods clearly outperform tightly coupled GFM-based methods, as they can operate directly on raw fisheye images and provide mature support for inertial measurements.

\subsection{Reconstruction Evaluation}
\label{sec:recon_eval}

To show the flexibility of our framework to the underlying SLAM system, we use OV-SLAM for \DATASETNAME\ dataset, and ORB-SLAM3 for Oxford Spires (mono-inertial) and KITTI (stereo) datasets.
Although not used in this section, our framework is also compatible with monocular SLAM systems.

\begin{figure}[t] 
    \centering    
    \includegraphics[
        width=\columnwidth,
        trim=0 0 0 5,  
        clip
    ]{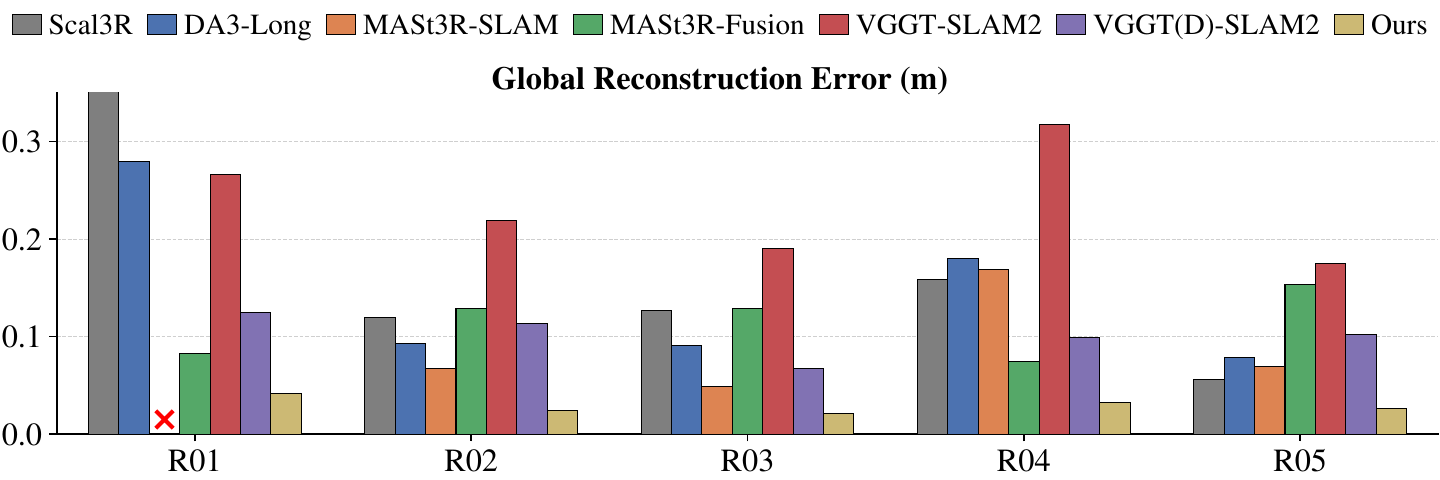}
    \caption{Global reconstruction error on the \DATASETNAME\ dataset.} 
    \label{fig:ori_global_cmp}
\end{figure}

Fig.~\ref{fig:ori_global_cmp} presents the global reconstruction error for different methods on \DATASETNAME\ dataset after aligning their trajectories and maps to ground-truth under $\mathrm{Sim}(3)$. Our method achieves the best performance by a wide margin---with an average error of $2.9\,\mathrm{cm}$, thanks to the decoupled design in which the underlying SLAM system yields more globally consistent trajectories.

\begin{table}[t!]
\centering
\scriptsize
\caption{Chunk-wise reconstruction quality on the \DATASETNAME\ dataset (chunk length $=10\,\mathrm{m}$, threshold $=3\,\mathrm{cm}$, conf. percentile $= 25\%$).}

\begin{tabular*}{\columnwidth}{@{\extracolsep{\fill}}lccccc}
\toprule
 & R01 & R02 & R03 & R04 & R05 \\
\midrule
\multicolumn{6}{c}{\textbf{Precision (\%) $\uparrow$}} \\[1pt]

Scal3R (Offline)
& 50.32
& 41.10
& 53.84
& 66.64
& 55.50 \\

DA3-Long 
& 65.64
& 50.32
& 62.52
& 55.46
& 56.58 \\

MASt3R-SLAM
& \gray{Failed}
& 58.80
& 64.01
& 54.25
& 64.61 \\

MASt3R-Fusion
& 62.97 
& 48.29 
& 52.16 
& 61.53 
& 54.36 \\

VGGT-SLAM2 
& 19.85 
& 20.19 
& 23.81 
& 26.16 
& 25.78 \\

VGGT(D)-SLAM2 
& 60.61
& 44.13
& 61.18
& 64.20
& 52.38 \\

Ours w/o Ext. 
& \underline{78.64} 
& \underline{76.41} 
& \underline{79.47} 
& \underline{80.50} 
& \underline{81.94} \\

Ours 
& \best{81.06} 
& \best{80.44} 
& \best{81.26} 
& \best{81.63} 
& \best{82.90} \\

\midrule
\multicolumn{6}{c}{\textbf{Reconstruction Error} \textbf{(m) $\downarrow$}} \\[2pt]

Scal3R (Offline)
& 0.0555
& 0.0886
& 0.0553
& 0.0365
& 0.0661 \\

DA3-Long 
& 0.0359
& 0.0582
& 0.0380
& 0.0470
& 0.0481 \\

MASt3R-SLAM 
& \gray{Failed}
& 0.0441
& 0.0378
& 0.0536
& 0.0372 \\

MASt3R-Fusion
& 0.0437 
& 0.0723 
& 0.0582 
& 0.0382 
& 0.0781 \\

VGGT-SLAM2 
& 0.1747 
& 0.1954 
& 0.1686 
& 0.1522 
& 0.1513 \\

VGGT(D)-SLAM2 
& 0.0391
& 0.0683
& 0.0409
& 0.0339
& 0.0582 \\

Ours w/o Ext. 
& \underline{0.0230} 
& \underline{0.0242} 
& \underline{0.0204} 
& \underline{0.0218} 
& \underline{0.0210} \\

Ours 
& \best{0.0216} 
& \best{0.0209} 
& \best{0.0198} 
& \best{0.0199} 
& \best{0.0198} \\

\midrule
\multicolumn{6}{c}{\textbf{Recall (\%) $\uparrow$}} \\[1pt]

Scal3R (Offline)
& 36.94
& 34.15
& 44.08
& 51.87
& 49.66 \\

DA3-Long 
& 51.84
& 38.34
& 51.41
& 46.28
& 51.42 \\

MASt3R-SLAM 
& \gray{Failed}
& 46.95
& 48.73
& 38.67
& 48.58 \\

MASt3R-Fusion
& 46.77 
& 36.96 
& 38.23 
& 44.52 
& 41.79 \\

VGGT-SLAM2 
& 21.91 
& 18.25 
& 18.12 
& 28.44 
& 27.77 \\

VGGT(D)-SLAM2 
& 40.08
& 31.82
& 50.12
& 52.07
& 53.08 \\

Ours w/o Ext. 
& \underline{53.97}
& \underline{50.74}
& \underline{54.31}
& \best{56.71}
& \best{57.92} \\

Ours 
& \best{57.72}
& \best{52.04}
& \best{55.38}
& \underline{55.84}
& \underline{56.24} \\

\bottomrule
\end{tabular*}

\vspace{3pt}
\raggedright
\footnotesize
\textit{Ours w/o Ext.}: pose inputs disabled for GFM prediction.

\label{tab:ori_recon_precision_cmp}
\end{table}

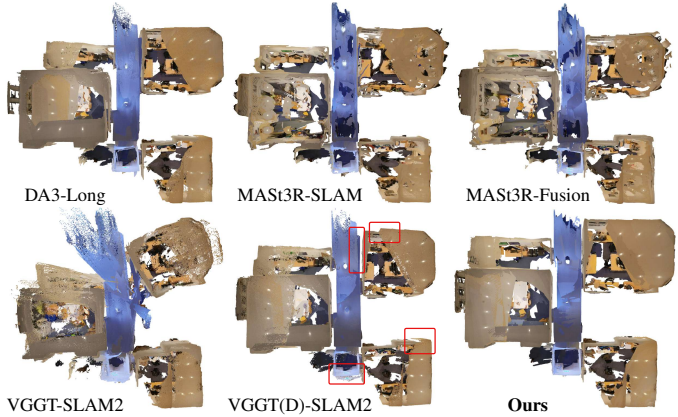
\begin{figure}[t!]
\centering
\begin{tikzpicture}
    \node[anchor=south west, inner sep=0] (img) at (0,0)
        {\includegraphics[width=\columnwidth]{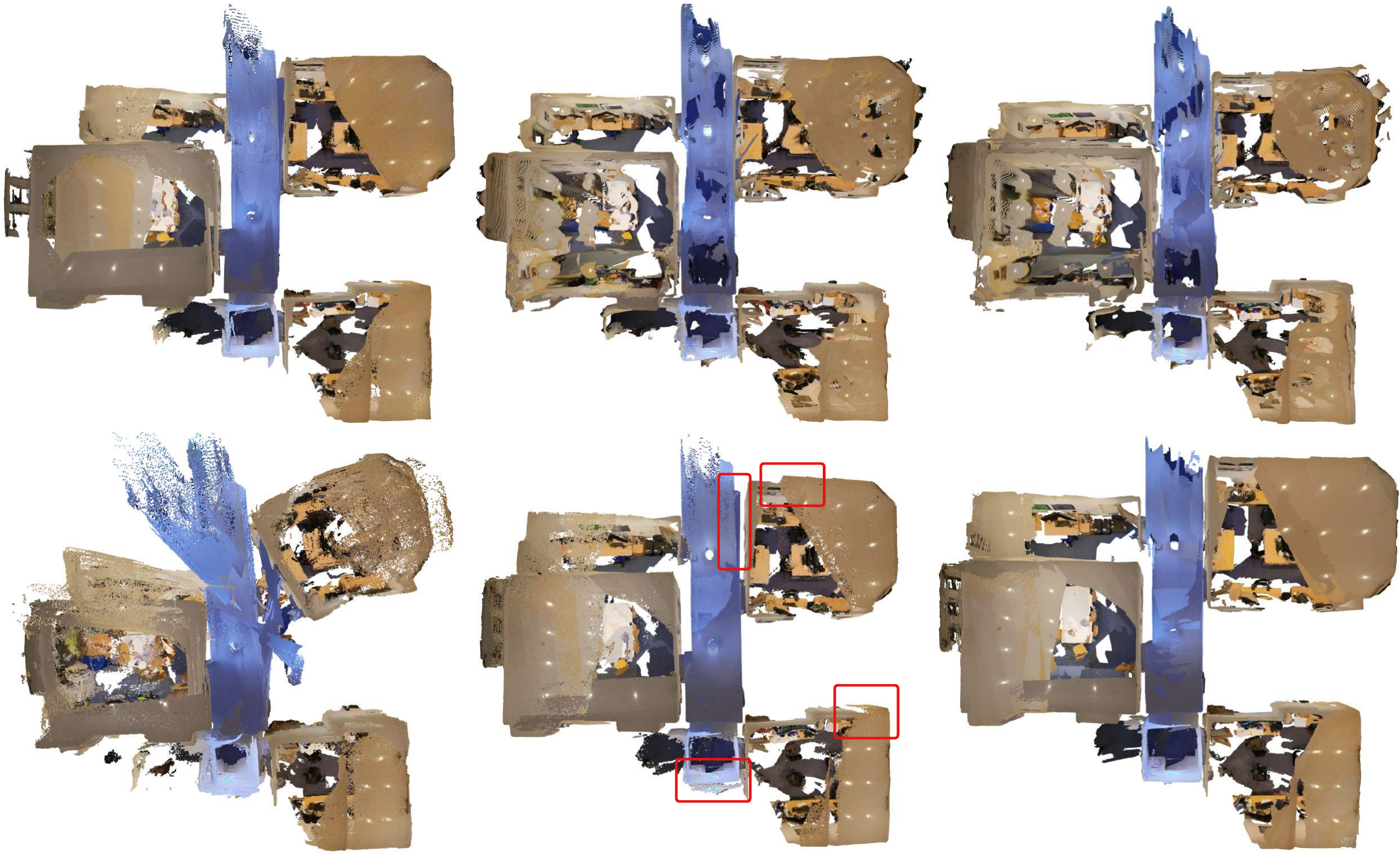}};
    
    \begin{scope}[x={(img.south east)}, y={(img.north west)}]
        \node[anchor=base, inner sep=0pt] at (0.09,0.51) {\scriptsize DA3-Long};
        \node[anchor=base, inner sep=0pt] at (0.44,0.51) {\scriptsize MASt3R-SLAM};
        \node[anchor=base, inner sep=0pt] at (0.78,0.51) {\scriptsize MASt3R-Fusion};
    
        \node[anchor=base, inner sep=0pt] at (0.09,0.00) {\scriptsize VGGT-SLAM2};
        \node[anchor=base, inner sep=0pt] at (0.44,0.00) {\scriptsize VGGT(D)-SLAM2};
        \node[anchor=base, inner sep=0pt] at (0.78,0.00) {\scriptsize \textbf{Ours}};
    \end{scope}
\end{tikzpicture}
\caption{Visualization of reconstruction on R03.}
\label{fig:r03_cmp_vis}
\end{figure}

To mitigate the influence of global trajectory quality and enable a more informative reconstruction comparison, we introduce chunk-wise evaluation.
Specifically, each trajectory is partitioned into chunks whose corresponding ground-truth trajectory length is $L$. For each chunk, the estimated trajectory is aligned to the ground-truth via a $\mathrm{Sim}(3)$ transformation, and the associated point cloud is transformed accordingly. We then apply an additional ICP refinement to register the reconstructed points to the ground-truth point cloud, from which the evaluation metrics are computed. The final performance is reported as the average over all chunks.

The mapping frame rate of our method is subsampled to $3\,\mathrm{Hz}$ for real-time operation; the same frequency is used for DA3-Long and VGGT(D)-SLAM2 to ensure a fair comparison. For MASt3R-SLAM and MASt3R-Fusion, we preserve their default frequency of $10\,\mathrm{Hz}$, as they use different GFM. We additionally include Scal3R~\cite{scal3r}, a purely offline reconstruction method, and evaluate it at $3\,\mathrm{Hz}$ since no default input image frequency is provided.

We apply 25th-percentile confidence filtering to all methods before evaluation. To ensure density consistency, we additionally perform voxel downsampling on the reconstruction, using a voxel size of $2\,\mathrm{cm}$ for the indoor dataset (\DATASETNAME) and $10\,\mathrm{cm}$ for outdoor datasets (KITTI and Oxford Spires).

\subsubsection{Indoor}
Due to the higher GPU memory demands of DA3-Long and VGGT(D)-SLAM2, DA3 can only process images with a batch size of 6 when used with these methods. For fairness, we use the same batch size for our method and also evaluate a variant without pose inputs to DA3 (w/o Ext.). For Scal3R, we use batch sizes of 16 to fit the memory constraints. Table~\ref{tab:ori_recon_precision_cmp} shows that our method outperforms the other approaches---including when operating without providing poses to DA3. This highlights the effectiveness of the scale optimization of our method. With pose inputs enabled (benefiting from the decoupled design), it outperforms other methods in precision by $10\%\text{--}20\%$, while maintaining around $2\,\mathrm{cm}$ reconstruction error. In terms of recall, the improvement is less pronounced as we apply 25th-percentile filtering, and recall is less sensitive to imperfect predictions when the same regions are observed from multiple viewpoints, allowing correct predictions to compensate for erroneous ones. For instance, double walling in GFM-based reconstructions can still achieve high recall despite degraded precision.

\begin{figure}[t] 
    \centering    
    \includegraphics[
        width=\columnwidth,
        trim=0 0 0 5,  
        clip
    ]{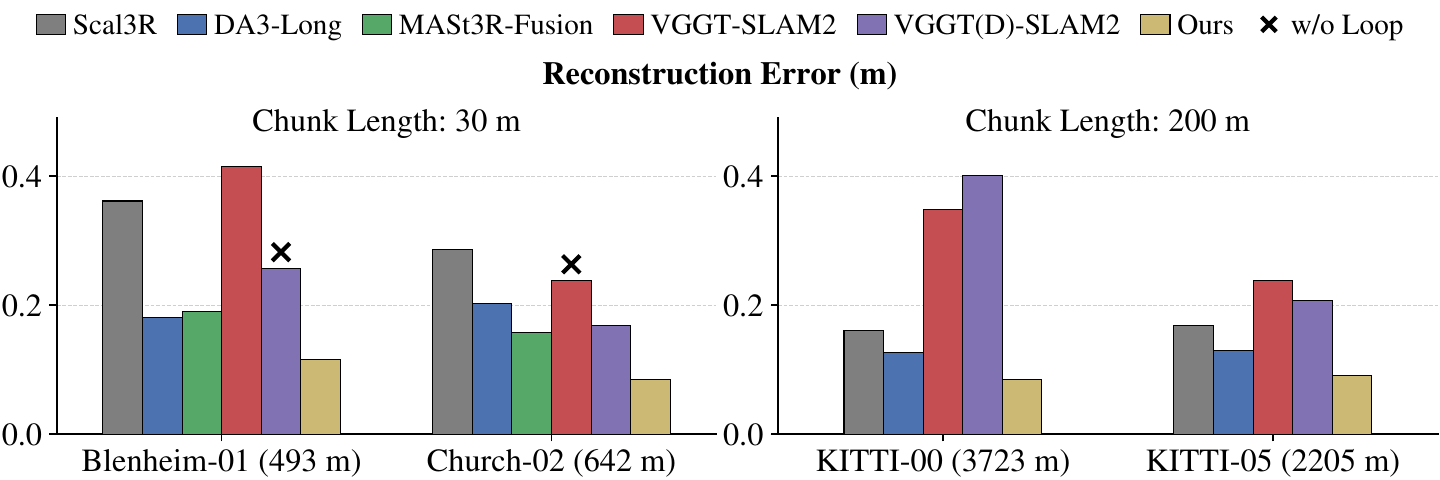}
    \caption{Chunk-wise reconstruction error on the outdoor datasets. After alignment to the ground-truth, points with depth greater than $20\,\mathrm{m}$ (in local coordinate) are discarded.} 
    \label{fig:outdoor_cmp}
\end{figure}

\begin{figure}[t] 
    \centering    
    \includegraphics[width=\columnwidth]{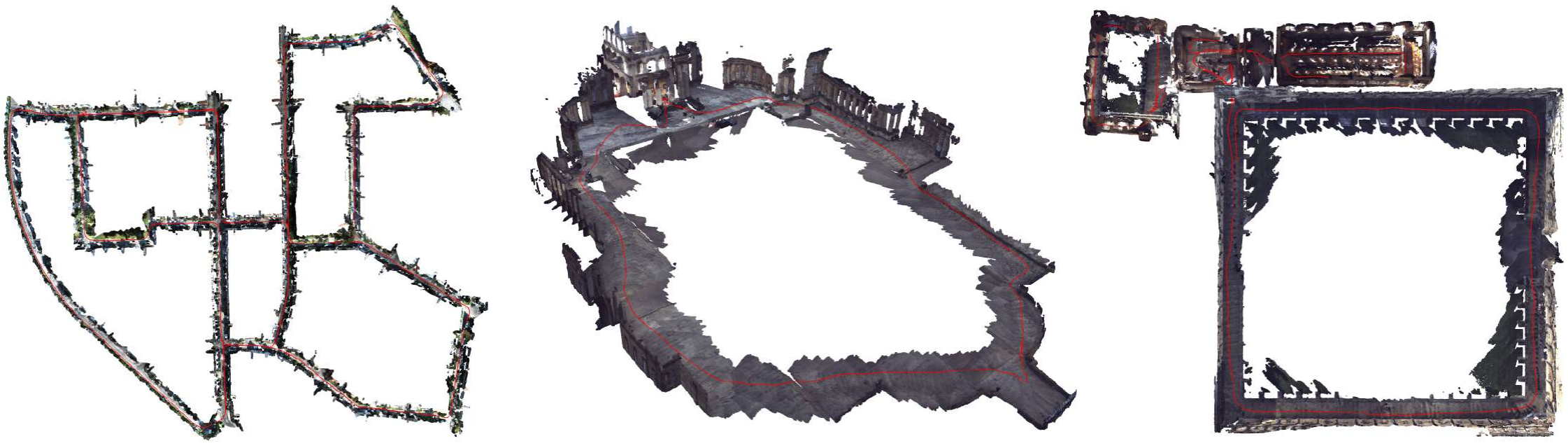}
    \caption{Outdoor reconstruction on KITTI-00, Blenheim-Palace-01, and Christ-Church-02.}
    \label{fig:outdoor_vis}
\end{figure}

Compared to VGGT(D)-SLAM2, which uses the same GFM and performs both inter- and intra-submap optimization on $\mathrm{SL}(4)$, our method simplifies the optimization to scale estimation only and anchors it to accurate poses provided by classical SLAM. This design mitigates drift and errors caused by the geometric inaccuracies in the GFM predictions.

\subsubsection{Outdoor}
For KITTI, the lower image aspect ratio allows larger batch sizes: 30 for DA3-Long and VGGT(D), and 60 for Scal3R. Our method continues to operate with a batch size of 6, demonstrating its robustness under limited batch-size. MASt3R-SLAM is excluded due to OOM issue, and MASt3R-Fusion is omitted on KITTI due to insufficient IMU data.

As shown in Fig.~\ref{fig:outdoor_cmp}, our method consistently achieves lower reconstruction error than competing approaches. We further observe that VGGT(D)-SLAM2 is prone to optimization failures in outdoor scenarios after loop closure, even when the detected loops are correct, despite operating reliably on our \DATASETNAME\ dataset (Fig.~\ref{fig:r03_cmp_vis}). Therefore, we revert it to odometry mode when global optimization fails (marked by \textbf{×}).

\begin{table}[t!]
\centering
\scriptsize
\caption{Ablation study on global reconstruction (batch size = 6).}

\begin{tabular*}{\columnwidth}{@{\extracolsep{\fill}}lccccc}
\toprule
 & R01 & R02 & R03 & R04 & R05 \\
\midrule

\multicolumn{6}{c}{\textbf{Global Precision (\%) $\uparrow$}} \\[1pt]

DepthAnything3 
& 78.91 
& 68.26 
& 72.00 
& 67.44 
& 70.16 \\

Ours w/o Frame Opt.
& \underline{82.52}
& \underline{81.87} 
& 80.73 
& 81.32 
& 77.40 \\

Ours w/o Submap Opt.
& \best{83.38} 
& 73.52 
& 77.18 
& 74.83 
& 76.31 \\

Ours w/o Points Fus.
& 81.30 
& 80.86 
& \underline{80.98} 
& \underline{83.96} 
& \underline{79.83} \\

Ours
& 81.81
& \best{81.93} 
& \best{81.73} 
& \best{84.88} 
& \best{80.97} \\

\arrayrulecolor{gray!60}
\midrule
\arrayrulecolor{black}

Imp. (abs. to DA3)
& +2.90
& +13.67
& +9.73
& +17.44
& +10.81 \\

Imp. (rel. to DA3)
& +3.7\%
& +20.0\%
& +13.5\%
& +25.9\%
& +15.4\% \\

\midrule

\multicolumn{6}{c}{\textbf{Global Reconstruction Error (m) $\downarrow$}} \\[2pt]

DepthAnything3 
& 0.0224 
& 0.0327 
& 0.0278 
& 0.0319 
& 0.0284 \\

Ours w/o Frame Opt.
& \best{0.0192}
& \underline{0.0199} 
& 0.0198 
& 0.0200 
& 0.0228 \\

Ours w/o Submap Opt.
& \underline{0.0197} 
& 0.0271 
& 0.0245 
& 0.0262 
& 0.0244 \\

Ours w/o Points Fus.
& 0.0201 
& 0.0203 
& \underline{0.0198} 
& \underline{0.0183} 
& \underline{0.0221} \\

Ours
& 0.0198
& \best{0.0198} 
& \best{0.0194} 
& \best{0.0178} 
& \best{0.0214} \\

\arrayrulecolor{gray!60}
\midrule
\arrayrulecolor{black}

Imp. (abs. to DA3)
& -0.0026 
& -0.0129 
& -0.0084 
& -0.0141 
& -0.0070 \\

Imp. (rel. to DA3)
& -11.6\% 
& -39.4\% 
& -30.2\% 
& -44.2\% 
& -24.6\% \\

\bottomrule
\end{tabular*}

\label{tab:ori_ablation_6}
\end{table}

\begin{table}[t!]
\centering
\scriptsize
\caption{Comparison with DA3 at batch size = 11.}

\begin{tabular*}{\columnwidth}{@{\extracolsep{\fill}}lccccc}
\toprule
 & R01 & R02 & R03 & R04 & R05 \\
\midrule

\multicolumn{6}{c}{\textbf{Global Precision (\%) $\uparrow$}} \\[1pt]
DepthAnything3 
& 83.73
& 77.56
& 80.51
& 79.10
& 76.02 \\

Ours
& \best{85.13}
& \best{82.64}
& \best{84.93}
& \best{85.28}
& \best{81.36} \\

\midrule

\multicolumn{6}{c}{\textbf{Global Reconstruction Error (m) $\downarrow$}} \\[2pt]
DepthAnything3 
& 0.0186
& 0.0220
& 0.0195
& 0.0216
& 0.0248 \\

Ours
& \best{0.0173}
& \best{0.0186}
& \best{0.0170}
& \best{0.0172}
& \best{0.0214} \\

\bottomrule
\end{tabular*}

\label{tab:ori_ablation_12}
\end{table}

\begin{figure}[t!] 
    \centering
    \includegraphics[width=0.48\columnwidth]{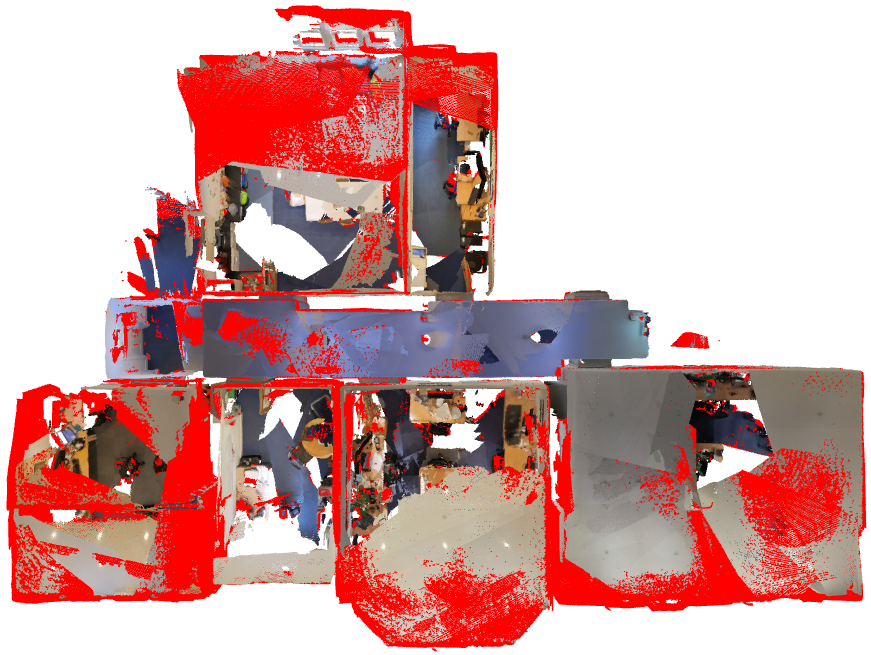}
    \hfill
    \includegraphics[width=0.48\columnwidth]{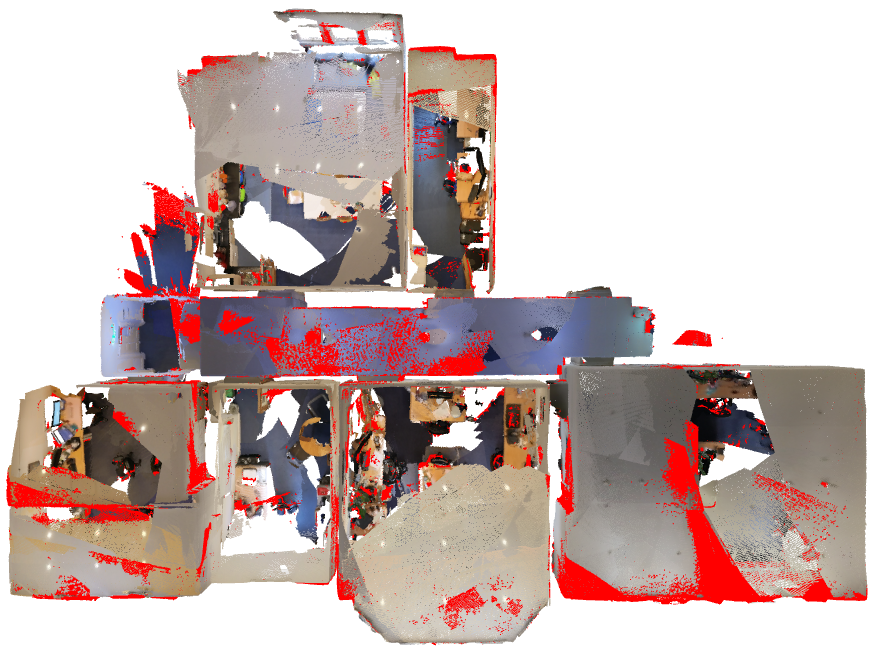}
    \caption{Comparison between DA3 reconstruction (left) and \METHODNAME\ (right) on R04, where red points indicate outliers.} 
    \label{fig:recon_cmp}
\end{figure}

\subsection{Ablation Study}
\label{sec:ablation}

Given that our system differs fundamentally from existing GFM-based SLAM systems, we also conduct a detailed ablation study to evaluate our proposed design. 
For all experiments, we use ground-truth poses as input; therefore, no chunk alignment is required, and reconstruction is evaluated globally.

\subsubsection{Scale Optimization}
As shown in Table~\ref{tab:ori_ablation_12} and~\ref{tab:ori_ablation_6}, smaller batch sizes reduce the spatial diversity of input images and degrade the geometric accuracy of DA3 predictions; therefore, we conduct ablation studies with a small batch size of 6 to better reveal the difference. Results in Table~\ref{tab:ori_ablation_6} show that our method significantly improves precision compared to directly aggregating DA3 predictions, achieving up to $17\%$ absolute improvement (Fig.~\ref{fig:recon_cmp}). We observe that both frame-level and submap-level scale optimization independently improve reconstruction quality, while disabling either component can lead to performance degradation, as evidenced in R04 and R05. Furthermore, point cloud fusion provides additional gains by reducing noise through confidence-weighted averaging.

Comparing Table~\ref{tab:ori_ablation_12} and Table~\ref{tab:ori_ablation_6}, we observe that our mapping framework is much less affected by reduced batch sizes compared to the direct use of DA3 ($-1.60\%$ and $-8.03\%$ in precision, respectively). This demonstrates the effectiveness of the proposed scale optimization. Notably, this characteristic is particularly important for devices with limited GPU memory or applications that require reactive mapping where the system must operate without waiting for a large number of incoming images, as is common in robotics domain.

\subsubsection{Keyframe Selection}
\begin{table}[t!]
\centering
\scriptsize
\caption{Comparison of reconstruction error ($\mathrm{m}$) with and without adaptive keyframe selection on the Oxford Spires dataset.}
\begin{tabular*}{\linewidth}{@{\extracolsep{\fill}}lcc}
\toprule
 & Blenheim-Palace-01 & Christ-Church-02 \\
\midrule
$\tau_t = 0.3\,\mathrm{m},\ d_{\max} = 20\,\mathrm{m}$ & 0.1468 & 0.1299 \\
adaptive ($\tau_t$, $d_{\max}$) & 0.1135 & 0.0863 \\
\bottomrule
\end{tabular*}
\label{tab:kf_selection_ablation}
\end{table}

Table~\ref{tab:kf_selection_ablation} compares our system with and without adaptive keyframe selection. Using a fixed threshold ($\tau_t = 0.3\,\mathrm{m}$) while keeping $d_{\max} = 20\,\mathrm{m}$ leads to higher reconstruction error due to an insufficient viewpoint baseline. In contrast, our keyframe selection strategy dynamically increases $\tau_t$ (capped at $1\,\mathrm{m}$) in far-range scenes and adjusts $d_{\max}$ based on the value of $\tau_t$ with $\alpha = 20$, thereby improving reconstruction quality.

\subsection{Runtime Analysis}
\label{sec:evaluation:runtime}
\begin{table}[t!]
\centering
\scriptsize

\caption{Runtime breakdown (sec) for batch size 6.}

\begin{tabularx}{\linewidth}{
>{\raggedright\arraybackslash}p{0.7cm}
*{5}{>{\centering\arraybackslash}X}
}
\toprule
& DA3 Infer. & LightGlue & Frm. Opt. & Points Fus. & Smp. Opt. \\
\midrule
Laptop & 0.9275 & 0.3196 & 0.0465 & 0.1014 & 0.1223  \\
Jetson & 4.6413 & 1.4319 & 0.2023 & 0.2571 & 0.4302  \\
\bottomrule
\end{tabularx}

\label{tab:runtime}
\end{table}

Table~\ref{tab:runtime} shows the runtime breakdown for processing a submap of 6 frames (including 1 overlapping frame), which takes $1.5\,\mathrm{s}$ on average. This corresponds to a maximum real-time keyframe rate of approximately $3.3\,\mathrm{Hz}$, assuming a SLAM system runs concurrently and provides posed frames asynchronously. In this paper, we set the keyframe frequency to $3\,\mathrm{Hz}$. We also evaluate the system on an NVIDIA Jetson AGX Orin platform, where runtime increases due to lower compute capability. In both cases, the main bottleneck is the inference of the deep learning models (DA3 and LightGlue).

A major computational cost arises from the sparse feature extraction and matching components. Our implementation is very loosely coupled with the underlying SLAM system, requiring only trajectory information. However, most modern visual SLAM systems already track matched features between frames and maintain a covisibility graph, and our method could directly leverage this information. Thus, a more tightly coupled implementation could significantly improve efficiency.

\subsection{Robot Experiment}
\definecolor{color_1}{RGB}{244, 179, 001}
\definecolor{color_2}{RGB}{219, 016, 072}

\begin{figure}[t] 
    \centering    

    \begin{tikzpicture}
        \node[inner sep=0] (img)
        {\includegraphics[width=\columnwidth]{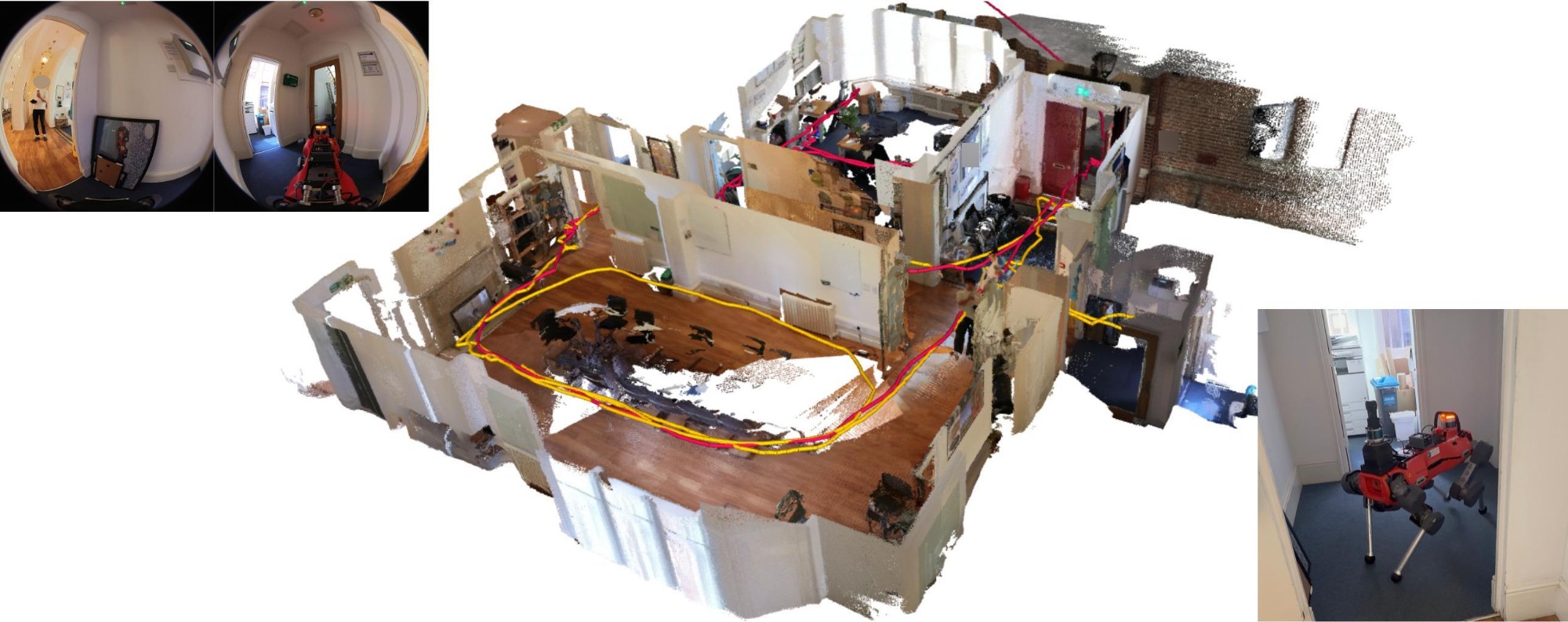}};

        \node[
            overlay,
            font=\small\bfseries,
            text=black,
            anchor=south west
        ] at ($(img.south west)+(0,0.5)$)
        {
            \tikz \fill[color_1] (0,0) rectangle (0.2,0.2);
            \hspace{0.3em}
            Session 1
        };

        \node[
            overlay,
            font=\small\bfseries,
            text=black,
            anchor=south west
        ] at ($(img.south west)+(0,0)$)
        {
            \tikz \fill[color_2] (0,0) rectangle (0.2,0.2);
            \hspace{0.3em}
            Session 2
        };

        \node[
            overlay,
            font=\scriptsize,
            text=black,
            anchor=south west,
            align=left
        ] at ($(img.south west)+(7.15,2.9)$)
        {
            Outdoor-Indoor\\Transition
        };
        
        \node[
            overlay,
            font=\scriptsize,
            text=black,
            anchor=south west,
            align=left
        ] at ($(img.south west)+(6.8,3.07)$)
        {
            ←
        };

    \end{tikzpicture}

    \caption{Multi-session, multi-camera, multi-modal dense SLAM on data collected by ANYmal-D with Insta360 ONE RS.} 
    \label{fig:robot_exp}
\end{figure}
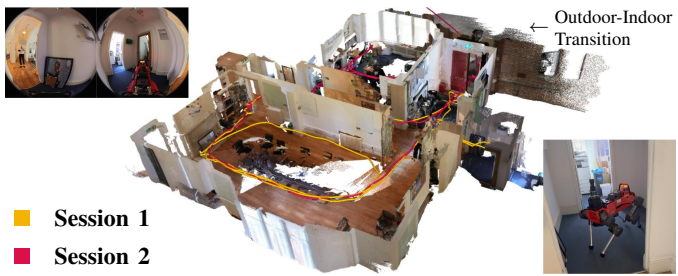

Our final experiment qualitatively evaluates ScaRF-SLAM using data collected by a quadruped robot (ANYmal-D) carrying an Insta360 camera, including an outdoor-indoor transition and operation in highly confined spaces. It showcases \textit{multi-session}, \textit{multi-camera} (front and rear), and \textit{multi-modal} (visual-inertial) SLAM with our system, highlighting the flexibility of the proposed decoupled framework and demonstrating its reliable operation in practical robotic scenarios (Fig.~\ref{fig:robot_exp}).

\section{Conclusion}
\label{sec:conclusion}

This work revisits the design choice of how GFMs should be incorporated into SLAM systems. Rather than pursuing tighter integration, we show that separating state estimation from dense reconstruction leads to a more robust and practical solution. Our results suggest that the limitations of current GFMs lie not in dense geometry generation, but in the reliability of their predictions for accurate pose estimation.
By anchoring reconstruction to more reliable poses from classical SLAM and by optimizing the scale of GFM dense point clouds, our proposed framework combines the strengths of both components. This simple design choice proves effective in practice, yielding consistent improvements in reconstruction. It also preserves accurate, low-latency tracking of the classical SLAM system and remains robust to small GFM input batch sizes.


Beyond the quantitative gains, our findings highlight a broader insight: progress in learned geometry does not necessarily require replacing classical geometric pipelines, but can instead be achieved through careful system level integration. In particular, enforcing lightweight geometric constraints on top of learned predictions provides a favorable balance between accuracy, efficiency, and robustness. We believe that such hybrid designs will play a key role in advancing dense mapping systems toward real-world deployment.


\balance
\small
\bibliographystyle{IEEEtran} 
\bibliography{string-short,references}

\end{document}